\documentclass[10pt,twocolumn,letterpaper]{article}
\usepackage{authblk}
 \usepackage[pagenumbers]{cvpr} 

\usepackage{graphicx}
\usepackage{amsmath}
\usepackage{amssymb}
\usepackage{booktabs}
\usepackage{booktabs}
\usepackage{tabularx}
\usepackage{multirow}
\newcolumntype{C}{>{\centering\arraybackslash}X} 
\usepackage{arydshln}
\usepackage{bbm}
\usepackage{enumitem}            
\usepackage{bm}
\usepackage{float}
\usepackage{xcolor}
\usepackage{todo}
\usepackage{balance}

\input{tbspace.tex}

\usepackage[pagebackref,breaklinks,colorlinks]{hyperref}

\newcommand{\secref}[1]{Section~\ref{sec:#1}} 
\newcommand{\figref}[1]{Figure~\ref{fig:#1}}  
 
\newcommand{\tabref}[1]{Table~\ref{tab:#1}}

\newcommand{\firstkey}[1]{\textcolor{blue}{\textbf{#1}}}

\usepackage{diagbox}
\usepackage{pifont}
\usepackage{xcolor}

\def\cvprPaperID{7032} 
\def\confName{CVPR}
\def\confYear{2022}

\usepackage[linewidth=1pt]{mdframed}
\mdfsetup{skipabove=0pt,skipbelow=0pt}

\begin{document}

\title{DaliID: Distortion-Adaptive Learned Invariance\\for Identification Models}
\author[1]{Wes Robbins$^*$}
\author[1,2]{Gabriel Bertocco$^*$}
\author[1]{Terrance E. Boult\thanks{This research is based upon work supported in part by the Office of the Director of National Intelligence (ODNI), Intelligence Advanced Research Projects Activity (IARPA), via [2022-21102100003]. The views and conclusions contained herein are those of the authors and should not be interpreted as necessarily representing the official policies, either expressed or implied, of ODNI, IARPA, or the U.S. Government. The U.S. Government is authorized to reproduce and distribute reprints for governmental purposes notwithstanding any copyright annotation therein.}}
\affil[1]{University of Colorado, Colorado Springs}
\affil[2]{Universidade Estadual de Campinas}
\affil[ ]{\tt\small wrobbins@uccs.edu, gabriel.bertocco@ic.unicamp.br, tboult@vast.uccs.edu}
\maketitle
\begin{abstract}In unconstrained scenarios, face recognition and person re-identification are subject to distortions such as motion blur, atmospheric turbulence, or upsampling artifacts. To improve robustness in these scenarios, we propose a methodology called Distortion-Adaptive Learned Invariance for Identification (DaliID) models. We contend that distortion augmentations, which degrade image quality, can be successfully leveraged to a greater degree than has been shown in the literature. Aided by an adaptive weighting schedule, a novel distortion augmentation is applied at severe levels during training. This training strategy increases feature-level invariance to distortions and decreases domain shift to unconstrained scenarios. At inference, we use a magnitude-weighted fusion of features from parallel models to retain robustness across the range of images. DaliID models achieve state-of-the-art (SOTA) for both face recognition and person re-identification on seven benchmark datasets, including IJB-S, TinyFace, DeepChange, and MSMT17. Additionally, we provide recaptured evaluation data at a distance of 750+ meters and further validate on real long-distance face imagery. 
\end{abstract}
\begin{figure}
    \centering
    \includegraphics[width=.44\textwidth]{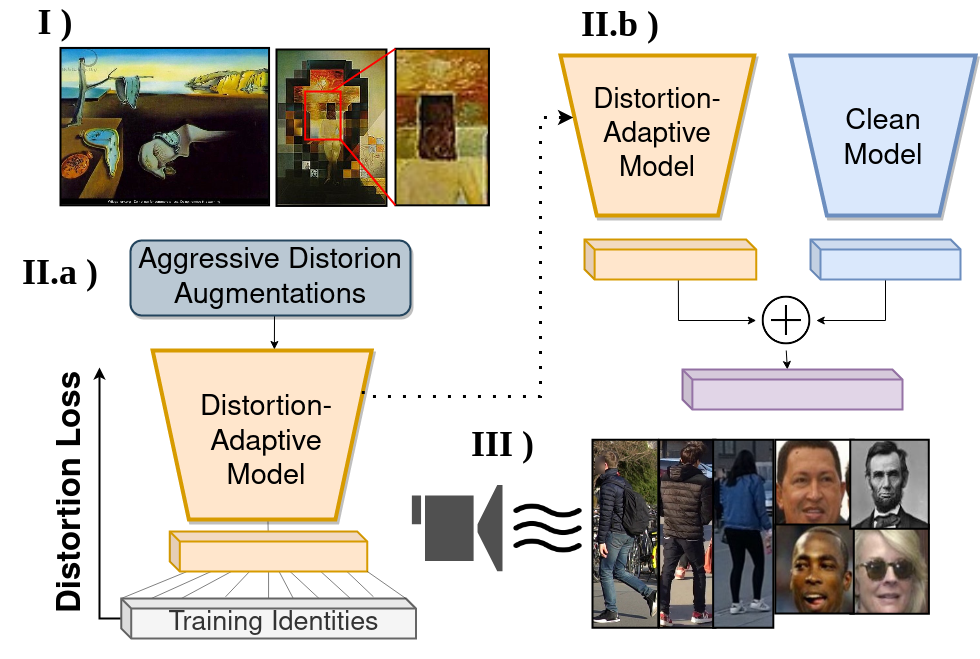} 
    \vspace{-12pt}
    \caption{\textbf{I)} Motivating the title of our method, works from renowned artist Salvador Dali leverage the human ability to see through distortions. To overcome realistic distortions encountered by in-the-wild biometric models, we propose \textbf{II.a)} a novel training procedure for distortion robust models and \textbf{II.b)} magnitude-weighted feature-fusion from high- and low-quality training domains. To supplement evaluations on realistic distortions, \textbf{III)} we collect and provide an IRB-approved academic-use dataset at a range of 750+ meters.}    
    \label{fig:front}
    
\end{figure}
\vspace{-8pt}
\section{Introduction}
\label{sec:intro}
The human visual system is capable of recognizing faces or objects before and after considerable distortions. 
Consider Dali's renowned works \textit{Persistence of Memory} and \textit{Lincoln in Dalivision} shown in part I of \figref{front} where the reader will have no trouble recognizing multiple clocks or Lincoln, despite the  distorted presentation. Comparatively, neural networks are brittle when presented with even mildly distorted images.  Within the field of biometrics, the tasks of face recognition and person re-identification can be subject to distortions at inference time, such as atmospheric turbulence,  motion blur, and artifacts from upsampling. Such distortions are common in security-sensitive settings such as energy infrastructure security, surveillance systems, or counter-terrorism~\cite{briar}. Thus, there is a significant social need for models that are robust in these conditions. 

To address robustness under challenging distortions, we propose \textbf{DaliID}: \textbf{D}istortion-\textbf{A}daptive \textbf{L}earned \textbf{I}nvariance for \textbf{Id}entification. 
DaliID is compromised of several novel contributions, which yield models with improved feature-level invariance to distortions and, therefore, are robust across evaluation scenarios. The DaliID methodology is demonstrated on the tasks of face recognition and person re-identification. The first insight of this work is that, for unconstrained conditions,  augmentations are under-utilized. We conjecture this is likely because simply adding degraded data as an augmentation adds so much noise to training that it degrades overall performance. While it is often accepted that cleaner data is better, we introduce  training and inference strategies that leverage distortion augmentations --- which degrade image quality --- and use them to improve  general performance. In particular, we propose an adaptive weighting mechanism during training to adjust the weights for each sample as a function of training iteration and ``distortion level." Images with higher distortion start the training with lower weighting, and images with lower distortion with greater weighting. The weighting of distorted samples is increased throughout training with a cosine scheduler.  Additionally, for person re-identification, we propose to use multiple class-centers and class-proxies that allow the model to better adapt to training distortions. The corresponding proxy loss (see \secref{adaptive_weighting} and the Supplementary) also follows the adaptive weighting schedule. Since face recognition has 100-1000x the number of training samples as person re-identification, multiple proxy class-centers are not necessary.

\vspace{-1mm}
To distort training samples, a novel \textit{distortion} augmentation is employed, which combines spatial distortion and blur. For the implementation of the augmentation, we leverage atmospheric turbulence simulation code~\cite{simulator}, allowing for physically realistic image distortions at various levels. With the adaptive weighting schedule, the augmentation is applied at severe levels. The range of distortion augmentations reduces the domain gap to challenging evaluation scenarios, which is significant for face recognition because the training data is web-scraped and is predominated by high-resolution celebrity images~\cite{webface}. Our distortion augmentation is effective to a surprising degree, and ablation studies show the advantage of our proposed distortion relative to Gaussian blur and down-sampling. The use of distortions during training implicitly supervises the model to learn a feature space that is invariant to such distortions and thus allows for improved generalization under such conditions. We refer to a model trained with the above procedure as a \textit{distortion-adaptive} model.

\vspace{-.7mm}
To further improve robustness at inference, two backbones are run in parallel: a distortion-adaptive backbone and a standard (or `clean') backbone. The final distance between samples for open-set evaluations is calculated with a magnitude-weighted combination of feature distances from each backbone respectively. Feature magnitude is used since it reflects the response of the learned features at the final layer which is known to be correlated with sample quality~\cite{Kim_2022_CVPR, magface, objectosphere}. Surprisingly, we find that this fusion approach is more robust than more complicated learned fusions such as an attention layer or full transformer encoder. Relative to a single distortion-adaptive backbone, the parallel backbone fusion improves performance on face recognition at low false-positive thresholds (e.g., IJB-C TAR@FAR=1e-4) and on all person re-identification benchmarks. The final result is a method that is highly robust across evaluation scenarios for both face recognition and person re-identification. The efficacy of DaliID is demonstrated empirically, including state-of-the-art performance on seven publicly available benchmarks:  IJB-S, IJB-C, TinyFace, CFP-FP, Market1501, MSMT17, and DeepChange. 

The final contribution of this work is the recapture of face recognition data over long-distance with high-end imaging equipment and displays. At 750+ meters, our proposed datasets have the longest range of any academic-use dataset available. The collection process and hardware specifications are discussed in detail in \secref{recollect} and in the Supplementary. In \secref{results}, prior works are compared on our proposed evaluation datasets. The datasets will be made available for academic use. The recapture of data is ongoing, and additional training data will be released before CVPR.

In summary, the contribution of this work includes:
\vspace{-.75em}
\begin{itemize}
    \setlength\itemsep{-.5em} 
    \item Propose a training augmentation based on atmospheric turbulence, which contains physically realistic spatial distortion and blur.
    \item Propose a novel distortion-adaptive training strategy in which we leverage the construction of distortion augmentation for an easy-to-hard weighting scheme. 
    \item Design a novel weighted combination strategy  based on the feature magnitudes from both backbones from the training phase, allowing us to exploit complementary knowledge and reach state-of-the-art performance across evaluation scenarios. 
    \item Provide identification datasets through long-distance (750+ meters) to  provide an assessment of the impact of significant atmospheric turbulence.
\end{itemize}


\begin{figure*}
    \centerline{\qquad
    \includegraphics[width=.93\textwidth]{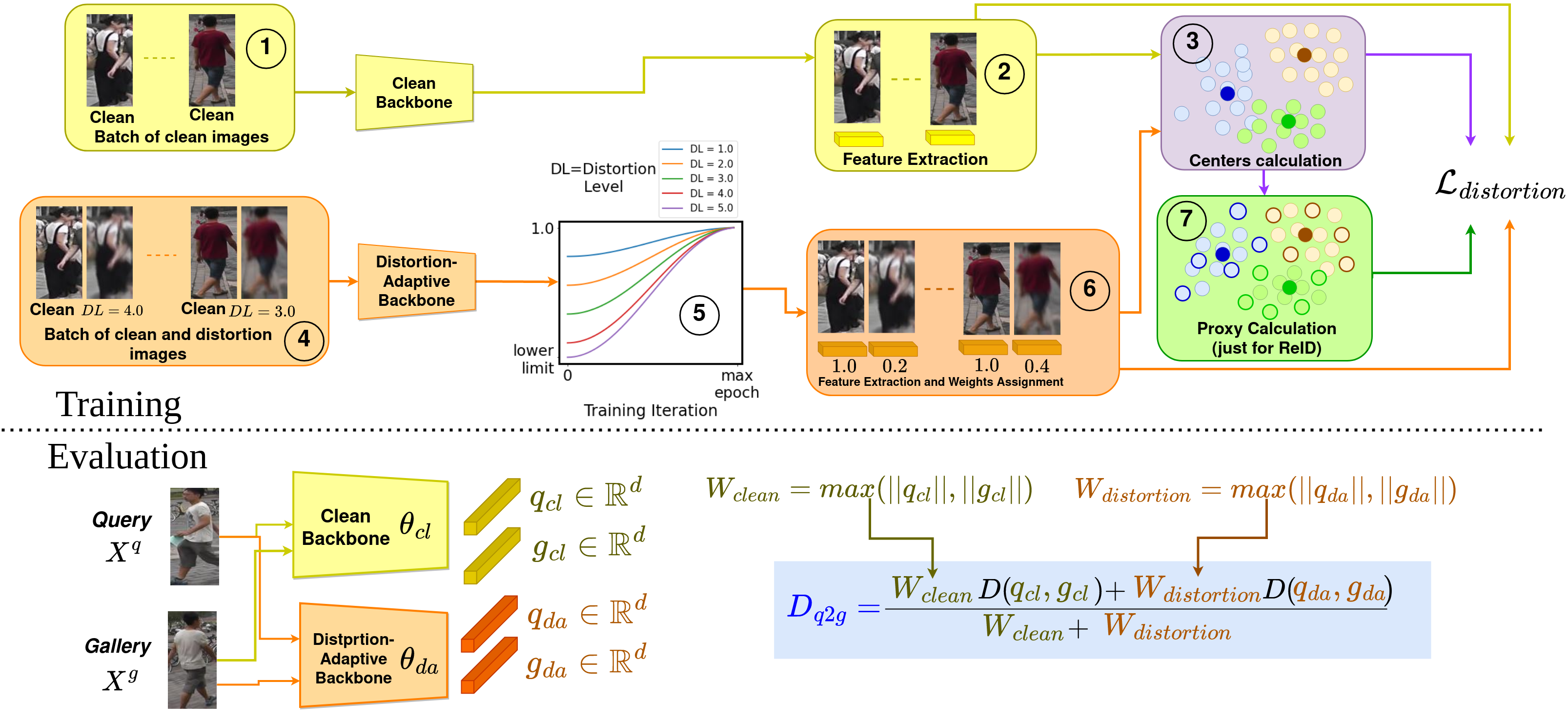}
    }
    \vspace{-4pt}
    \caption{An overview of the DaliID pipeline for face recognition (DaliFace) and person re-identification (DaliReID). Steps 1,2 and 3 are performed for training without distorted images, while steps 3,4,5,6 are distortion-adaptive training. In Step 4, we create a batch of clean and distorted images, then a dynamically varied weight is assigned as a function of the distortion level (\textit{DL}) (Step 5). Then we extract the features and optimize the distortion loss ($\mathcal{L}_{distortion}$). Step 7 is applied just for DaliReID training due to the high intra-class variation to sample count ratio faced in the whole-body recogntion task. On evaluation, both clean and distortion-adaptive backbone decisions are weighted and combined based on the magnitudes of the query and gallery feature vectors to obtain the final decision (distance) for retrieval.\vspace{-10pt}}
    \label{fig:pipeline_overview}
\end{figure*}
\section{Related Work}
The problems of face recognition and person re-identification have been extensively studied. Most related to this paper are works that have studied low quality conditions. For face recognition, PFE~\cite{pfe} proposes representing faces with a Gaussian distribution in latent space to account for uncertainty. Data Uncertainty Learning (DUL)~\cite{dul} builds on PFE by learning the mean and variance of the Gaussian distribution during training. URL~\cite{url} uses data synthesis and a confidence-aware loss to learn universal representations. Several quality-aware face recognition loss functions have also been proposed. CirriculurFace~\cite{curriculurface} changes the margin of the loss throughout training and the MagFace~\cite{magface} and AdaFace~\cite{Kim_2022_CVPR} losses use adaptive margins that are a function of feature magnitude, which is a proxy for quality. CFSM~\cite{cfsm} is a method that learns the style of a test environment and uses a latent style model to modify training samples. CAFace is a clustering-based method for multi-frame face recognition~\cite{kim2022cluster}. In \cite{Robbins_2022_CVPR}, the effects of atmospheric turbulence on face recognition are studied, where atmospheric distortions are found to significantly affect face recognition performance. Other works have developed upstream image restoration for atmospheric turbulence~\cite{yasarla_cnn_based, yasarla_uncertain, atfacegan}.  Image restoration methods focus on image-based metrics such as PNSR, not recognition.

For Person Re-Identification (PReID), CBDB-Net~\cite{tan2021incomplete} proposes the Batch DropBlock to encourage the model to focus in complementary parts of the input image. CDNet~\cite{li2021combined} improves architecture search for PReID. FIDI~\cite{yan2021beyond} proposes a novel loss function to give different penalizations based on distances between images to encourage fine-grained feature learning. To deal with clothes-changing, CAL~\cite{gu2022clothes} regularizes the model learning with respect to the clothes labels to learn clothes-invariant features. There are many other prior art that leverage attention models~\cite{chen2019abd,zhang2021person,fang2019bilinear,ye2021deep,hou2019interaction,chen2019mixed,rao2021counterfactual,zhao2020not,zhang2020relation,huang2022avpl,wang2022nformer,zhu2022dual,li2021diverse} neighborhood-based analysis~\cite{wang2022nformer}, auxiliary data~\cite{he2021transreid, jia2022learning}, segmentation-based~\cite{kalayeh2018human}, semantics-based~\cite{jin2020semantics} and part-based learning~\cite{sun2018beyond,zheng2019pyramidal,wang2020high,wang2018learning,wang2022feature,zhu2022learning,zhang2021person,ye2021deep,zhu2020identity,zhu2022learning}.
To directly deal with different resolutions and points of view, some works leverage the camera information associated to each identity~\cite{zhuang2020rethinking}, super-resolution strategies~\cite{jiao2018deep, cheng2020inter}, and attention and multi-level mechanisms for cross-resolution feature alignment~\cite{zhang2021deep, munir2021resolution}. There is insufficient space to compare orthogonally to all combinations of described methods above for PReID. We limit our scope comparison to global feature representation learning model as described in the taxonomy of the recent survey from Ye et al.~\cite{ye2021deep}, in which we just perform global pooling operations over the last feature map of a CNN without further mechanisms. The core contributions of this paper are focused on learning distortion-invariant feature-spaces and a methodology for dealing with distortion, which is demonstrated to be applicable to both face recognition and person re-identification. Future work should look at combining techniques such as image-restoration, super-resolution, part-based mechanisms, or multi-frame aggregation with the improved feature spaces developed herein.

\section{Approach}
\label{sec:approach}


We propose DaliID for learning models robust to realistic test-time distortions such as motion blur, upsampling artifacts, and atmospheric turbulence. We use strong levels of distortion augmentation (\secref{aug}), which serves the purpose of supervising the model to learn a feature space that is invariant to distortions that have been shown to considerably degrade model performance~\cite{yasarla_uncertain, Robbins_2022_CVPR}. 
To allow the model to adapt to strong levels of augmentation, we propose an adaptive-weighting distortion-aware strategy (\secref{adaptive_weighting}) where we dynamically change the weights of different distortion levels throughout training. 
To get the highest performance across the range of evaluation scenarios, we train two models in parallel: one with clean images and the other with clean and distorted images (\secref{cross}). Then, we perform a weighted combination of the feature spaces from both models based on the magnitude of the feature vectors from each, which yields the highest performance. DaliID methodology is designed for general identification scenarios such as face recognition and person re-identification tasks. An overview of the approach is shown in Figure~\ref{fig:pipeline_overview}.
\subsection{Distortion Augmentations}
\label{sec:aug}
Image augmentations allow better generalization by adding variance to training data. There is a vast space of augmentations that can be performed on an image; many have been successful for computer vision tasks. However, there is a bias-variance trade-off. In this work, we leverage a new augmentation for face recognition and PReID training based on atmospheric turbulence to generate the different distortion levels for the images. Atmospheric turbulence contains random temporally and spatially variable distortions, which are not present in Gaussian blur or down-sampling augmentations.  Atmospheric turbulence simulation code~\cite{simulator} is used to implement the augmentation, which generates physically realistic distortions. Our approach of simulated distortions is of practical interest because it is not tractable to collect real labeled data through atmospherics at a scale suitable for training deep learning models. Experimentally, we find training with our distortion augmentation yields state-of-the-art performance on long-distance and low-resolution test sets. Distortion levels used herein are based on different atmospheric turbulence conditions to train our models. 

\subsection{Adaptive Weighting}
\label{sec:adaptive_weighting}
Different levels of distortion compress different degrees of difficulty during training. 
Randomly sampling images from different distortion levels can result in sub-optimal performance since higher distortion levels (i.e., lower-quality samples) dominate the gradient during training.
In counterpart, hard-training mining strategies have shown promising performance in PReID models~\cite{hermans2017defense, sheng2020mining} and face recognition~\cite{curriculurface}. In this context, we propose an easy-to-hard training regime in which we start by assigning higher weights for lower levels of distortion and lower weights for higher levels of distortion. \textit{Different than prior works, we directly leverage the construction of the augmentation to assign weights.} Weighting the loss as a function of the distortion level allows the model to focus on easier examples (by giving them higher weights). By lower the weighting of high-distortion samples, the model becomes distortion-aware without allowing them to dominate the loss in early epochs. As the training progresses, the weights for all distortion levels increase according to a cosine schedule. 
An illustration of the weighting for each distortion level is shown in Step 5 of Figure~\ref{fig:pipeline_overview} and is formally described below.


The distortion-aware training considers a batch of images $B = \{X^{i}\}_{i=1}^{N_{b}}$ that is composed by a mix of clean images $X_{cl}^{i}$ and distorted images $X_{dl}^{i}$ with distortion level randomly sampled from five possible values ($dl \in \{1,2,3,4,5\}$), where $N_{b}$ is the batch size. A higher $dl$ value indicates a stronger distortion. Then features $f_{t}^{i}$, with $t \in \{cl,dl\}$ are extracted from the backbone ($\theta_{da}$). During the loss calculation, the respective weight $w_{t}^{i}$ is assigned to each image according to the cosine weighting schedule. These steps are shown in Steps 4, 5, and 6 in Figure~\ref{fig:pipeline_overview}. For the same distortion level, the weights increase along the training following a cosine schedule (Step 5). After that, the centers are obtained for each class (Step 3), and, if we are performing PReID training, we also take the classes' proxies in Step 7. The distortion loss is calculated as follows:
\begin{equation}
\label{eq:ce_loss}
\begin{aligned}
    \mathcal{L}_{ce}(f,q,P)= &&\\
    -log&\frac{e^{cos(\omega_{fq}+m_1)/\tau)+m_2}}{e^{cos(\omega_{fq}+m_1)/\tau)+m_2}+ \sum\limits_{\substack{p \in P, p\neq q}}e^{(cos\omega_{fp})/\tau}}
\end{aligned}
\end{equation}
\vspace*{-1ex}
\begin{equation}
\displaystyle
\label{eq:center_loss}
    \mathcal{L}_{distortion} = \frac{1}{W}\sum_{i=1}^{N_{b}}\sum_{t \in \{cl, dl\}}w_{t}^{i}\mathcal{L}_{ce}(f_{t}^{i}, p_{+}, P)
\end{equation}

\noindent where $p_{+}$ is the positive class-center (i.e., proxy), $P$ is the set of all class-centers, $\omega_{fq}$ is the angle between vectors $f$ and $q$ (same definition for $\omega_{fp}$), and $W = \sum_{i=1}^{|B|}\sum_{t \in \{cl, dl\}}w_{t}^{i}$. For hyperparameters, $\tau$ is temperature, $m_1$ is angular margin, $m_2$ is the additive margin. For face recognition $\tau = 1$ and $m_1,m_2$ are adaptive as proposed in AdaFace~\cite{Kim_2022_CVPR}. For PReID, $m_1,m_2 = 0$ and $\tau=0.05$. Further implementation details are in the supplementary.


In PReID, to better adapt to distortions, we extend the use of multiples proxies\cite{wang2021camera} to the supervised case. This is necessary due to limited training samples and high intra-class variance, which occurs since the whole-body images are captured from  different cameras resulting in views of the same person in different poses, illumination conditions, backgrounds, occlusions, and resolutions leading to high intra-class variance and low inter-class distances~\cite{wang2020high, ye2021deep}. Step 7 of Figure~\ref{fig:pipeline_overview} shows the multiple proxies with the circles with dark outlines. For the sample $X^{i} \in B$, the proxies of the same class are denoted by $P_{i}$, and the top-50 closest negative proxies (from other classes) to $X^{i}$ are denoted by $N_{i}$. Then we define the proxy loss as:
\vspace{-8pt}
\begin{equation}
\label{eq:proxy_loss}
    \begin{aligned} 
    & \mathcal{L}_{proxy} = \\ &\qquad \frac{1}{W}\sum_{i=1}^{N_{b}}\sum_{t \in \{cl, dl\}}w_{t}^{i}\frac{1}{|P_{i}|}\sum_{q \in P_{i}}L_{ce}(f_{t}^{i}, q, P_{i} \cup N_{i}).
    \end{aligned}
\end{equation}

\setlength\dashlinedash{0.3pt}
\setlength\dashlinegap{1.5pt}
\setlength\arrayrulewidth{0.4pt}

\begin{table*}
\centering
\resizebox{1.0\textwidth}{!}{
\begin{tabular}{l|c|cc|cccc|cccc|cc|cc}
\hline\hline
 \multirow{2}{*}{Method} &  \multirow{2}{*}{Dataset}  & \multicolumn{2}{c|}{TinyFace~\cite{tinyface}}    & \multicolumn{4}{c|}{IJB-S S-to-B~\cite{ijbs}}    & \multicolumn{4}{c|}{IJB-S S-to-S~\cite{ijbs}}  & \multicolumn{2}{c|}{LFW-LD (Sec. \ref{sec:recollect})} &  \multicolumn{2}{c}{CFP-LD (Sec. \ref{sec:recollect})}  \\
  
  &  & \multicolumn{1}{c}{Rank-$1$} & \multicolumn{1}{c|}{Rank-$5$}  & \multicolumn{1}{c}{Rank-$1$} & \multicolumn{1}{c}{Rank-$5$} & \multicolumn{1}{c}{$1\%$} & \multicolumn{1}{c|}{$10\%$} & \multicolumn{1}{c}{Rank-$1$} & \multicolumn{1}{c}{Rank-$5$} & \multicolumn{1}{c}{$1\%$} & \multicolumn{1}{c|}{$10\%$} & \multicolumn{1}{c}{C-to-LD} & \multicolumn{1}{c|}{LD-to-LD} & \multicolumn{1}{c}{C-to-LD} & \multicolumn{1}{c}{LD-to-LD}\\
  \hline 
PFE~\cite{pfe}\scalebox{0.875}{\textcolor{white}{aaa}}  & MS1Mv2~\cite{arcface}   & -  & - & $53.60$ & $61.75$  & $35.99$ & $39.82$ & $9.20$  & $20.82$   & $0.84$ & $2.83$ & -  & - &-&-   \\
ArcFace~\cite{arcface}  & MS1Mv2~\cite{arcface}    & - & - & $57.36$ & $64.95$    & $41.23$ & -  & - &- & - & -  & - & - &-&-  \\
URL~\cite{url}   & MS1Mv2~\cite{arcface}  & $63.89$ &  $68.67$  & $61.98$ & $67.12$    & $42.73$ & -    & - & -    & - & -&-&-  \\
CF~\cite{curriculurface}  & MS1Mv2~\cite{arcface}    & $63.68$ & $67.65$ & $63.81$ & $69.74$    & $47.57$ & - &  $19.54$ & $32.80$   & $2.53$ & -  & -& -&-&- \\
AdaFace~\cite{Kim_2022_CVPR}   & MS1Mv2~\cite{arcface}   &  $68.21$ & $71.54$  & $66.27$ & $71.61$    & $50.87$ & - & $23.74$ & $37.47$ & $2.50$  & -  &  - & - &-&-  \\
\hline 
ArcFace~\cite{arcface} & WF4M~\cite{webface}  &  $71.11$  &  $74.38$& $68.38$ & $73.64$ & $52.47$ & $60.69$ & $27.20$ & $38.36$ & $\textbf{4.30}$ & $\firstkey{15.95}$& $87.80$ & $84.65$ & $78.12$ & $71.17$ \\
AdaFace~\cite{Kim_2022_CVPR} &  WF4M~\cite{webface}   &  $\firstkey{72.02}$  &  $\firstkey{74.52}$ & $\firstkey{69.52}$ & $\firstkey{74.41}$ & $\textbf{54.92}$ & $\firstkey{62.82}$ & $\firstkey{27.90}$ & $\firstkey{40.11}$ & $4.20$ & $14.44$& $\firstkey{89.20}$ & $\firstkey{86.10}$ & $\firstkey{79.87}$ & $\firstkey{72.57}$  \\
\textbf{DaliFace (ours)} & WF4M~\cite{webface} & $\textbf{73.98}$ & $\textbf{77.07}$ & $\textbf{72.21}$ & $\textbf{76.77}$ & $\firstkey{54.07}$ & $\textbf{63.10}$ & $\textbf{30.65}$ & $\textbf{42.33}$ & $\firstkey{4.21}$ & $\textbf{16.73}$ & $\textbf{93.91}$ & $\textbf{88.15}$ & $\textbf{83.21}$ & $\textbf{74.61}$\\
\hline
AdaFace~\cite{Kim_2022_CVPR} &  WF12M~\cite{webface}  &  $72.29$  &  $74.52$ & $69.73$ & $74.49$ & $\textbf{56.86}$ & $\firstkey{63.98}$ & $28.83$ & $40.99$ & $\textbf{4.04}$ & $15.11$ &  $89.89$ & $86.32$ & $80.57$ & $71.71$\\
CFSM~\cite{cfsm} & WF12M~\cite{webface} & $\firstkey{73.87}$ & $\firstkey{76.77}$ & $\firstkey{70.36}$ & $\firstkey{75.89}$ & $55.92$ & $63.63$ & $\firstkey{30.44}$ & $\firstkey{41.57}$ & $3.78$ & $\firstkey{15.88}$& $\firstkey{90.88}$ & $\firstkey{86.62}$ & $\firstkey{83.13}$ & $\textbf{75.10}$ \\
\textbf{DaliFace (ours)} &  WF12M~\cite{webface} & $\textbf{74.76}$ & $\textbf{77.36}$ & $\textbf{72.19}$ & $\textbf{76.66}$ & $\firstkey{56.04}$ & $\textbf{64.37}$ & $\textbf{32.25}$ & $\textbf{43.03}$ & $\firstkey{3.81}$ & $\textbf{16.97}$& $\textbf{94.00}$ & $\textbf{89.10}$ & $\textbf{83.98}$ & $\firstkey{74.96}$\\
\hline\hline
\end{tabular}
    }

\vspace{-2mm}
\caption{Comparison of DaliFace to prior work on benchmarks containing distortions. Common metrics are reported for TinyFace~\cite{tinyface} and for IJB-S$^{**}$~\cite{ijbs} protocols surveillance-to-booking (S-to-B) and surveillance-to-surveillance (S-to-S). For LFW-LD and CFP-LD (\secref{recollect}), 1:1 verification accuracy with clean-to-long-distance (C-to-LD) pairs and long-distance-to-long-distance (LD-to-LD) pairs. KEYS. \textbf{Bold}: First; \firstkey{Blue}: Second. $^{**}$ The IJB-S dataset contains over 3 million raw video frames and 15 million face annotations. Face recognition results are subject to detection and pre-processing steps. Using official code and pre-trained model, all WebFace\{4M,12M\} comparisons are run with our pre-processing to ensure fair comparison. See the supplementary for extensive detail on IJB-S evaluation.\vspace{-3mm}}
\label{tab:low_qual_face}
\end{table*}
The class proxies $P_{i}$ in equation~\ref{eq:proxy_loss} are defined such that each one is furthest from the others to get maximum class coverage. 
The final loss function for PReID is defined as: 
\vspace{-8pt}
\begin{equation}
\displaystyle
\label{eq:distortion_loss}
    \mathcal{L}_{distortion} = \mathcal{L}_{center} + \lambda \mathcal{L}_{proxy}
\end{equation}
\noindent where $\lambda$ controls the contribution of $\mathcal{L}_{proxy}$ to the final loss. $\mathcal{L}_{distortion}$ is applied for both distortion-adaptive and clean backbones training. To train the clean backbone, we have $w_{i} = 1$ for all samples because no distortion augmentations are applied. Following prior work for PReID, we employ the Mean-Teacher~\cite{tarvainen2017mean} strategy, which performs a self-ensembling of the weights parameters along training and has been employed in prior works. Further implementation details are in the Supplementary.  

\subsection{Cross-Domain Fusion}
\label{sec:cross}
The distortion-adaptive backbone improves performance on face recognition benchmarks and in person re-identification, but not on all high-quality scenarios for face recognition at low false-positive thresholds. In practice, we do not know the test-time distortion level, and thus a good model should be robust across all scenarios. We train a backbone without distortion augmentations, denoted $\theta_{cl}$, in parallel to the distortion-adaptive backbone, denoted $\theta_{da}$. To leverage knowledge from both backbones, we apply magnitude-weighted fusion between the backbones as shown in Figure~\ref{fig:pipeline_overview}. We call this cross-domain fusion since the backbones were trained on different training distributions. The advantage of this approach is evident in \tabref{ablation_weighting}.
At inference, for a query and gallery image pair, we extract both feature vectors $q_{cl} = \theta_{cl}(X^{q})$ and $g_{cl} = \theta_{cl}(X^{g})$ from query and gallery images pair considering the clean model with parameters $\theta_{cl}$, and the feature vectors $q_{da} = \theta_{da}(X^{q})$ and $g_{da} = \theta_{da}(X^{g})$. We calculate the distance between the query and gallery considering each backbone to obtain distances $D(q_{cl}, g_{cl})$ and $D(q_{da},g_{da})$, which are weighted combined considering the maximum feature magnitude for each pair before L2 normalization as shown in the equation on the lower half of Figure~\ref{fig:pipeline_overview}.

\begin{figure}[b!]
    \centering
    \includegraphics[width=.9\linewidth]{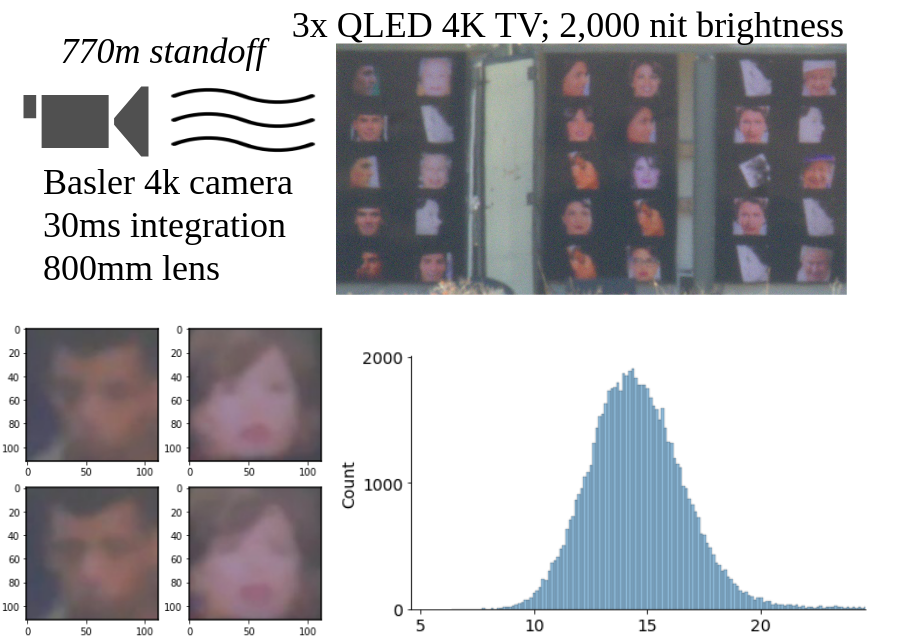}
    \caption{\textbf{Top.} Recapture specifications and a raw frame from our recollection. \textbf{Lower Left.} Two consecutive frames (33.3ms apart) for two different identities from LFW-LD. Differences can be observed between sequential frames, such as around the eyes or face outline. \textbf{Lower Right.} Distribution of feature distances in degrees between sequential frames of the same display image from LFW-LD and CFP-LD. Surprisingly, the distances are not 0 -- the effects of atmospherics from frame to frame are considerable! \vspace{-5pt}}
    \label{fig:recap_fig}
\end{figure}
\vspace{-10pt}
\vspace{8pt}
\section{Long Distance Recapture Data}
\label{sec:recollect}
As discussed in \secref{intro}, long-range recognition is relevant in many applications. However, the collection of biometric data is extremely expensive and time-consuming. Currently, the most related dataset, IJB-S~\cite{ijbs}, is not available for common academic use, and an earlier dataset at 100M \cite{gunther2017unconstrained} was withdrawn from public use. Furthermore, IJB-S is not a strictly long-range dataset. To overcome the lack of available data, prior works have used simulated atmospheric turbulence as a proxy for real data~\cite{yasarla_cnn_based, yasarla_uncertain, Robbins_2022_CVPR}. However, the efficacy of simulated atmospherics for face recognition has not been validated because, as mentioned before, there is no real data for validation.

To facilitate academic research on biometric recognition over long distances, we recapture datasets through the atmosphere. To perform the capture, we use three 4k outdoor televisions, a 4k Basler camera, and an 800mm lens with a 1.4x adapter. Custom capture and display software are developed for the collection, and custom mounting hardware is built for stable capture. The displays are mounted to avoid direct sunlight on the screens. The camera is directed at the displays from a structure at a distance of 770 meters, and videos of the displays are captured at 30 frames per second. A video of the displays running is provided in the supplementary material, where considerable atmospheric effects can be seen. Our collection setup yields significant atmospheric distortions, which can even be noticed between sequential frames. \figref{recap_fig} shows two examples. Details on the collection setup are in the Supplementary. (The collection of the data went through IRB approval.)

We refer to our recapture datasets as the original dataset name followed by ``-LD," (``the LD datasets"), where LD stands for long-distance. The evaluation datasets provided are LFW-LD and CFP-LD. Twelve recaptured samples for are provided for each image in the original dataset because atmospheric turbulence is temporally variable. For CFP-LD and LFW-LD, two protocols are proposed: clean-to-long-distance (C-to-LD) and long-distance-to-long-distance (LD-to-LD). C-to-LD uses verification pairs where one image is standard (and thus higher quality). For, LD-to-LD, all samples are recaptured over long distances.  The LD datasets supplement the evaluation of our methods in the following section, where evaluations are made with a single frame for each image. However, future work should perform research with frame fusion or frame selection across frames. Previous unconstrained evaluation datasets (e.g., IJB-S) have been distributed as over a terabyte of raw video, which is burdensome to process. In contrast, the LD datasets are pre-processed and pre-aligned in the same format as the original datasets, which streamlines evaluation and comparison. The final  release for CVPR will  include the recapture of person re-identifications datasets, plus a WebFace4M recapture for training.

\section{Results}
\label{sec:results}
Our experiments are performed on the tasks of face recognition and person re-identification with an emphasis on low-image-quality scenarios. Common training and evaluation procedures are followed for each task respectively.
\subsection{Datasets}
\label{sec:datasets}
We evaluate the face recognition models on five low-image-quality datasets and four standard image-quality datasets and then on recaptured and real-long distance data.  The low-resolution TinyFace~\cite{tinyface} dataset has 2,569 probe identities and 157,871 gallery images. Following previous work~\cite{Kim_2022_CVPR}, 1:N Rank 1 and Rank 5 are presented for TinyFace. The IJB-S~\cite{ijbs} contains gallery images for 201 identities and over 30 hours of probe video. For IJB-S, we report the surveillance-to-booking and surveillance-to-surveillance protocols. Additional details on IJB-S evaluation can be found in the Supplementary. Our LFW-LD and CFP-LD datasets (see \secref{recollect}) are evaluated with 1:1 accuracy. Representing standard image quality scenarios, we report on LFW~\cite{lfw}, CFP~\cite{cfp}, AgeDB~\cite{agedb}, and IJB-C~\cite{ijbc} with standard metrics. Training is done on the WebFace4M and WebFace12M datasets~\cite{webface}. Training is not performed on MS1Mv* datasets due to redaction.  

For person re-identification, we used two same-clothes datasets: Market1501 and MSMT17, and one clothes-changing dataset: DeepChange. For PReID evaluation, following prior work, experiments are run with predefined train-test splits and mAP and CMC metrics are reported. Market1501~\cite{zheng2015scalable} has 12,936 images of 751 identities in the training set. 
The test set is divided into 3,368 images for the query set and 15,913 images for the gallery set. 
MSMT17~\cite{wei2018person} is the most challenging same-clothes ReID dataset. It comprises 32,621 images of 1,401 identities in the training set and 93,820 images of 3,060 identities in the test set. 
DeepChange~\cite{xu2021deepchange} has lower-quality images than MSMT17 and Market. It has 75,083 images of 450 identities on the training set. The validation and test sets are divided into query and gallery sets. 

\vspace{-5mm}
\paragraph{Real Long-Distance (RLD)  Dataset.}
In addition to the previous public data, a non-public government-owned long-distance identification dataset is used for added validation.  For evaluation, we use a gallery  of 375 subjects with one high-quality image for enrollment. These are compared with 1,219 probe images captured at multiple distances up to 500m. See supplemental files for more details and image examples.   Performance on the RLD provides another comparison under real  atmospheric turbulence and a comparison to our LFW-LD and CFP-LD. 
\vspace{-1mm}
\subsection{Experimental Settings}
\label{sec:experiment_settings}
Common experimental settings are used for face recognition and person re-identification, respectively. For face recognition, a ResNet100~\cite{he2016deep} is used as the backbone model with an embedding size of 512. Mixed precision floating point training ~\cite{micikevicius2018mixed} is used, and the total batch size is 1,024. Stochastic Gradient Descent (SGD) is used as the optimizer with polynomial weight decay of $5e^{-4}$, and momentum of $0.9$. A base learning rate $0.1$ is used with a polynomial learning rate scheduler. In addition to distortions augmentations discussed in \secref{aug}, a horizontal flip and crop are used as augmentations. 

For fair comparison to the prior person re-identification work, we adopt the ResNet50~\cite{he2016deep} as the model backbone. Following previous works~\cite{luo2019strong, munir2021resolution}, we change the stride of the last residual block to 1 to increase the feature map size. Then we insert a global average pooling and global max pooling layer after the last feature map and sum their outputs element-wise~\cite{munir2021resolution}. After that, we add batch normalization and perform the L2-normalization to project them to the unit hyper-sphere. Further implementation details are in the Supplementary material.
    
\subsection{Comparison to state-of-the-art methods}
\label{sec:hyper}



\begin{table}[th]
\centering

\resizebox{1.0\linewidth}{!}{
\begin{tabular}{ l| c  c  c  c }
\hline\hline
Method & LFW   & CFP-FP & AgeDB &  IJB-C  \\ 
\hline 
\multicolumn{5}{c}{MS1Mv* Training} \\
\hline
CosFace \cite{cosface}                        (CVPR18)& 99.81 & 98.12  & 98.11 & 96.37  \\ 
ArcFace  \cite{arcface}                       (CVPR19)   & 99.83 & 98.27  & 98.28  & 96.03  \\    
GroupFace  \cite{groupface}            (CVPR20) & 99.85 & 98.63  & 98.28  & 96.26  \\      
CircleLoss \cite{circleloss}               (CVPR20)  & 99.73 & 96.02  & -     & 93.95  \\  
DUL \cite{dul}                                (CVPR20)   & 99.83 & 98.78  & -    & 94.61  \\
CF \cite{curriculurface}                      (CVPR20)   & 99.80 & 98.37  & 98.32  & 96.10  \\
URL \cite{url}                                (CVPR20)  & 99.78 & 98.64  & -    & 96.60  \\
DB \cite{Cao_2020_CVPR}                       (CVPR20)   & 99.78 & -      & 97.90   & -      \\
Sub-center \cite{subcenter}                 (ECCV20)   & 99.80 & 98.80  & 98.31 & 96.28  \\ 
BroadFace \cite{broadface}             (ECCV20)   & 99.85 & 98.63  & 98.38 & 96.38  \\   
VPL \cite{vpl}                (CVPR21)   & 99.83 & 99.11  & 98.60  & 96.76  \\ 
VirFace \cite{virface}                  (CVPR21) & 99.56 & 97.15  & -  & 90.54  \\  
DCQ  \cite{dcq}                         (CVPR21)  & 99.80 & 98.44  & 98.23   & -      \\  
MagFace~\cite{magface}                        (CVPR21)   & 99.83  & 98.46   & 98.17 & 95.97 \\
Virtual FC \cite{virtualfc}               (CVPR21)   & 99.38 & 95.55  & -     &  71.47  \\ 
CFSM~\cite{cfsm}                              (ECCV22)   & - & - & - & 95.90 \\
\hline 
\multicolumn{5}{c}{WebFace4M Training} \\
\hline
ArcFace~\cite{arcface}                        (CVPR19) & \firstkey{99.83} & 99.19 & \firstkey{97.95} & 97.16 \\
AdaFace~\cite{Kim_2022_CVPR}                  (CVPR22) & 99.80 & 99.17 & 97.90 & \firstkey{97.39} \\
Partial FC~\cite{partialfc}                   (CVPR22) & \textbf{99.85} & \firstkey{99.23}  & \textbf{98.01} & 97.22  \\ 
\textbf{DaliFace (ours)}                               & \firstkey{99.83} & \textbf{99.27} & 97.85 & \textbf{97.40}\\
\hline\hline
\end{tabular}
}
\vspace{-2mm}
\caption{\label{tab:compare}Performance comparisons between DaliFace and prior works on relatively high image quality scenario benchmarks. Following the most common protocols, 1:1 verification accuracy is reported for LFW~\cite{lfw}, CFP-FP~\cite{cfp}, and AgeDB~\cite{agedb}; and TAR@FAR=1e-4 is reported for IJB-C~\cite{ijbc}. Despite our focus on low quality scenarios (see \tabref{low_qual_face}, \tabref{rld_table}), it can be seen that our models are competitive with or better than state-of-the-art models on popular high image quality benchmarks. \textit{{\color{darkgray} Due to redaction, we do not perform training with MS1Mv1,2,3 datasets.\vspace{-3mm}}}}
\end{table}

\paragraph{Face Recognition} In \tabref{low_qual_face}, DaliFace is compared to prior works on low-image-quality benchmarks. In the WebFace4M regime, DaliFace improves over the prior state-of-the-art on TinyFace by 1.96\% on Rank-1 and 2.55\% on Rank-5. On IJB-S, DaliFace achieves state-of-the-art on 6 out of 8 metrics by an average of 1.53\%. On the long-distance datasets (LFW-LD and CFP-LD), DaliFace averages 3.04\% higher accuracy than prior work. In the WebFace12M regime, the prior SOTA is CFSM~\cite{cfsm}, which uses a latent-style model to learn the domain of the testing data. In contrast, our work does not use the testing data, but still improves over CFSM on 8/8 IJB-S metrics and 2/2 TinyFace metrics. In total, DaliFace achieves SOTA in 12/14 metrics in the WebFace4M regime and 11/14 metrics in the WebFace12M regime.

\tabref{compare} shows results on high-quality datasets IJB-C, LFW, AgeDB, and CFP. Despite using significant distortions during training, our DaliFace methodology has comparable or higher performance on high-quality benchmarks. SOTA is reached on IJB-C with an accuracy 97.40\% and on CFP-FP with an accuracy of 97.27\%.
\begin{table}[b]
    \centering
    \resizebox{1.0\linewidth}{!}{
    \begin{tabular}{l|c c c c}
    \hline\hline
    \multicolumn{5}{c}{\textit{Real Long-Distance (RLD) dataset}} \\
    \hline
     Method    &  Rank-1 & Rank-5 & 1\% & 10\% \\
     \hline
     ArcFace~\cite{arcface}  & $47.42$ & $57.26$ & $55.95$ & $69.32$ \\
     MagFace~\cite{magface}  & $45.69$ & $57.67$ & $56.52$ & $69.73$\\
     AdaFace~\cite{Kim_2022_CVPR}  & $49.96$ & $59.15$ & $58.24$ & $71.21$ \\
     \hdashline
     Distortion Aug (ours) & $56.52$ & $66.28$ & $62.10$ & $75.72$\\
     Distortion-Adaptive (ours) & $56.77$ & $\textbf{67.27}$ & $63.17$ & $\textbf{76.13}$ \\
     DaliFace (ours) & $\textbf{56.93}$ & $67.02$  & $\textbf{63.67}$ & $75.98$\\
     \hline\hline
    \end{tabular}
    }
    \vspace{-3mm}
    \caption{Performance on the Real Long-Distance (RLD) dataset, which contains real images captured at up to 500 meters (see \secref{datasets} for details). All models are trained on WebFace4M~\cite{webface}. The improvement of  DaliFace over other prior state-of-the-art algorithms is more than the gaps between previous algorithms and is consistent with other experiments.}
    \label{tab:rld_table}
\end{table}

To show that the improvements hold for actual long-distance data, we also compared DaliFace to various algorithms on the Real Long Distance dataset, with the results in \tabref{rld_table}.  It can be seen that our algorithm significantly improves over prior works across metrics on real long-distance data. DaliFace achieves TPR@FPR of 63.7\% @  1\%;  the next best algorithm is AdaFace at 58.3\%.  

\vspace{-13pt}
\paragraph{Person Re-identification}
DaliReID is compared with state-of-the-art methods in PReID for both the same-clothes scenario and the clothes-changing scenario.
For the same-clothes scenario, results are reported in Table~\ref{tab:state_of_art_same_clothes}. 
Our method is orthogonal to the backbone, and we show results with two backbones used in prior works: ResNet50 and OSNet~\cite{zhou2019omni}. On the Market dataset, we achieve the highest performance, outperforming FIDI~\cite{yan2021beyond} by $0.8$ in mAP, and the second position (along with FIDI) with R1 = $94.5$. In MSMT17, the most challenging PReID benchmark, we reach the best performance by outperforming CDNet by a margin of $5.9$ and $3.2$ in mAP and R1, respectively, with ResNet50. With OSNet, we achieve the best performance in both datasets for both metrics. Our method is able to rank ground-truth gallery images closer to the query and outperforms prior art in mAP in all setups.
To show our model generalization ability, we trained DaliReID for DeepChange, in which subjects' clothes differ among views. The results in Table~\ref{tab:state_of_art_changing_clothes} show our method also outperformed the state-of-the-art methods. 
We outperformed the recent CAL~\cite{gu2022clothes} by $2.9$ and $6.8$ in mAP and R1, respectively. Besides the clothes changing, DeepChange has more distortions and low-quality data than Market and MSMT17, and we obtain the highest gain on it for R1 and the second highest gain for mAP (after MSMT17), showing our method can better improve performance in low-quality datasets. We do not employ any kind of part-based, alignment, segmentation mask, nor pose variation strategies, in order to verify the performance improvement brought just by our DaliReID model. 

\begin{table}
    \scriptsize
    \centering
    \resizebox{1.0\linewidth}{!}{
    \begin{tabular}{ l | c | c c | c c}
    \hline\hline
    \multicolumn{2}{c}{} &
    \multicolumn{2}{|c|}{\textbf{Market}} &
    \multicolumn{2}{c}{\textbf{MSMT17}} \\ \hline
    Method & Venue & mAP & R1 & mAP & R1 \\
    \hline
    \multicolumn{6}{c}{\textit{OSNet-based models}} \\ \hline
    OSNet~\cite{zhou2019omni} & {\footnotesize ICCV19} & 84.9 & 94.8 & 52.9 & 78.7 \\
    \textbf{DaliReID (OSNet)} & \textbf{This work} & \textbf{87.2} & \textbf{95.0} & \textbf{59.5} & \textbf{82.6} \\ \hline
    \multicolumn{6}{c}{\textit{ResNet50-based models}} \\ \hline
    GCS~\cite{chen2018group} & CVPR18 & 81.6 & 93.5 & - & -\\
    SFT~\cite{luo2019spectral} & ICCV19 & 82.7 & 93.4 & 47.6 & 73.6\\
    CBN~\cite{zhuang2020rethinking} & ECCV20 & 83.6 & 94.3 & - & -\\
    STNReID~\cite{luo2020stnreid} & {\footnotesize TMM20} & 84.9 & 93.8 & - & - \\
    CBDB-Net~\cite{tan2021incomplete} & TCSVT21 & 85.0 & 94.4 & - & - \\
    BAT-Net~\cite{fang2019bilinear} & ICCV19 & 85.5 & 94.1 & 50.4 & 74.1 \\
    CDNet(*)~\cite{li2021combined} & {\footnotesize CVPR21} & 86.0 & \textbf{95.1} & \firstkey{54.7} & \firstkey{78.9} \\
    FIDI~\cite{yan2021beyond} & {\footnotesize TMM21} & \firstkey{86.8} & \firstkey{94.5} & - & - \\ 
    \hline
    \textbf{DaliReID (R50)} & \textbf{This work} & \textbf{87.6} & \firstkey{94.5} & \textbf{60.6} & \textbf{82.1} \\
     \hline\hline
    \end{tabular}
    }
    \vspace{-8pt}
    \caption{Comparison to the state-of-the-art models in same-clothes Person Re-Identification setup. \textbf{Bold} and \firstkey{Blue} indicate the best and second-best values. *CD-Net is not based on ResNet50 but the authors of the paper mostly compared to ResNet50-based models so we leave it here for fair comparison. \vspace{-2mm}}
    \label{tab:state_of_art_same_clothes}
\end{table}

\begin{table}[b!]
    \footnotesize
    \centering
    \resizebox{.9\linewidth}{!}{
    \begin{tabular}{ l | c | c c }
    \hline\hline
    \multicolumn{2}{c}{} &
    \multicolumn{2}{c}{\textbf{DeepChange}} \\ \hline
    Method & Venue & mAP & R1 \\
    \hline
    ReIDCaps~\cite{huang2019beyond} & {\footnotesize TCSVT20} & 11.3 & 39.5\\
    ViT~\cite{xu2021deepchange} & {\footnotesize ArXiv20} & 15.0 & 49.8\\
    ViT (with Grayscale)~\cite{xu2021deepchange} & {\footnotesize ArXiv20} & 15.2 & 48.0\\
    CAL~\cite{gu2022clothes} & CVPR22 & \firstkey{19.0} & \firstkey{54.0}\\
    \hline
    \textbf{DaliReID (R50)} & \textbf{This work} & \textbf{21.9} & \textbf{60.8} \\
    \hline\hline
    \end{tabular}
    }
    \vspace{-8pt}
    \caption{Comparison to the state-of-the-art models in clothes-changing person re-identification setup. \textbf{Bold} and \firstkey{Blue }indicates the best sencond-best values. All methods, except ViT, consider ResNet50 (R50) as backbone.}
    \label{tab:state_of_art_changing_clothes}
\end{table}
\vspace{-1mm}
\subsection{Ablation Study}
\paragraph{Face Recognition}
To demonstrate the improvements of the respective components of DaliFace, \tabref{ablation_weighting} shows an ablation with datasets representing three different evaluation scenarios: IJB-S for standard-quality, CFP-LD for long-distance, and TinyFace for low spatial resolution. It can be seen that aggressive distortion augmentations create a significant performance improvement in low quality datasets CFP-LD and TinyFace, however, performance drops significantly on IJB-C, which is a dataset with relatively higher quality images. After adding adaptive weighing and cross-domain fusion, it can be seen that the final model (i.e., DaliFace) is the best performing. In \tabref{rld_table}, which also ablates different components of our model, a similar pattern can be observed. An additional ablation of our distortion augmentation compared to Gaussian blur and down-sampling is provided in the supplementary.

\vspace{-15pt}
\paragraph{Person Re-identification}
\begin{table}[ht]
\footnotesize
\begin{center}
\fbox{
\parbox{.93\columnwidth}{
\footnotesize
\resizebox{1.0\linewidth}{!}{
    \begin{tabular}{ l | c c c | c}
    \hline\hline
    Face Ablation  &  IJB-C & CFP-LD & TinyFace & Average \\
    \hline
    Baseline ($\theta_{cl}$)  & 97.38 & 75.22 & 72.18 & 81.59\\
    Distortion Aug& 96.91 & 78.16 & 74.11 & 83.06\\
    Distortion-Adaptive ($\theta_{da}$) & 96.92& 78.37 & \textbf{74.22} & 83.17 \\
    \textbf{DaliFace} & \textbf{97.40} & \textbf{78.91} & 73.98 & \textbf{83.43}\\ [1ex]
     \hline\hline
\end{tabular}\\
}
\resizebox{1.0\linewidth}{!}{
\centering
\begin{tabular}{l | c c  c c cc}
\multicolumn{1}{c|}{} &
\multicolumn{2}{c}{Market} &
\multicolumn{2}{c}{MSMT17} &
\multicolumn{2}{c}{DeepChange} \\ \hline
 ReID Ablation  & mAP & R1 & mAP & R1 & mAP & R1\\ \hline
Baseline ($\theta_{cl}$) & 86.6 & 94.2 & 57.6 & 80.3 & 20.5 & 59.3\\
Distortion Aug & 86.3 & \textbf{94.7} & 55.4 & 78.5 & 20.2& 58.6 \\
Distortion-Adaptive (no $\mathcal{L}_{proxy}$) & 82.4 & 92.9 & 47.9 & 72.9 & 19.2 & 55.6\\
Distortion-Adaptive ($\theta_{da}$)& 86.6 & 94.3 & 58.3 & 81.3 & 20.7 & 59.2 \\
\textbf{DaliReID} & \textbf{87.6} & 94.5 & \textbf{60.6} & \textbf{82.1} & \textbf{21.9} & \textbf{60.8} \\
\hline
\end{tabular}}}
}
\end{center}
\vspace{-15pt}
\caption{Ablation study for both face and PReID datasets. The respective first lines shows performance of a baseline model. The second and third lines are for backbones trained with distortion as augmentation and our adaptive weighting strategy respectively. For PReID, line 4 ablates the proxy loss (all other PReID lines contain $\mathcal{L}_{proxy}$), and the final line is the proposed DaliReID model. CFP-LD is reported as an average of the two protocols shown in \tabref{low_qual_face}.\vspace{-2mm}}
\label{tab:ablation_weighting}
\end{table}
We perform a set of ablation studies over the PReID datasets to measure the impact of different components. In \tabref{ablation_weighting}, we ablate the different components of DaliReID. When we use distorted images as augmentations without our adaptive-weighting strategy (second line) we see a performance drop for both metrics in MSMT17 and DeepChange, and for mAP in Market when compared to the distortion-adaptive and DaliReID models. For MSMT17 and DeepChange, the results are also worse than the clean model showing that just employing distortion as augmentations hinders model performance. We face the same performance dropping when we take out our proxy loss ($\lambda = 0$ in Eq.~\ref{eq:distortion_loss}), showing it is an essential contribution (third line of Table~\ref{tab:ablation_weighting}). In contrast, just the distortion-adaptive backbone (fourth line) yields performance improvements for both MSMT17 and DeepChange, and for mAP in Market showing that it can learn a distortion-invariant feature space to some extent. Our final DaliReID model combines both clean and distortion-adaptive backbones (first and fourth lines), which leads to the best performance for MSMT17 (an increase of $5.2$ and $3.6$ for mAP and R1 respectively) and DeepChange (an increase of $1.7$ and $2.2$ for mAP and R1 respectively). This shows that DaliReID can effectively combine knowledge from both backbones.
In the future we aim to apply our methodology in PReID datasets considering moving cameras (e.g., UAV) with distortion levels caused by distance and altitude~\cite{kumar2020p}. 
\vspace{-1mm}
\section{Conclusion}
In this work, DaliID is presented as a methodology for improving robustness to distortions that are common in real-world applications. The proposed components include distortion augmentation, distortion-adaptive weighting, and a parallel-backbone magnitude-weighted feature fusion. While face recognition and person re-identification have considerable differences, DaliID is shown to be applicable in both tasks with state-of-the-art performance on seven datasets. The proposed LD datasets allow for further evaluation of realistic distortions and are captured over the longest distance of any academic dataset. 
\vspace{-14pt}
\paragraph{Limitations} A limitation of this work is that two parallel backbones are used for the final DaliID models. Since the backbones are run in parallel, inference time is not affected -- if there is sufficient hardware for the increased computational overhead. Future work should explore weight sharing between the backbones. 

\vspace{-12pt}
\paragraph{Potential Societal Impact}
Biometric models have a positive social impact on authentication and security applications, however, misuse of the technology can result in privacy infringements. For our research, we strictly adhere to licensing guidelines and do not use redacted datasets, and we encourage the community to avoid the use of redacted datasets, such as MS1Mv*. 
\balance
{\small
\bibliographystyle{ieee_fullname}
\bibliography{refs}
}

\end{document}


\title{DaliID: Distortion Adaptation and Learned Invariance\\for Deep Identification Models - Supplementary Material}

\author[1]{Wes Robbins$^*$}
\author[1,2]{Gabriel Bertocco$^*$}
\author[1]{Terrance E. Boult\thanks{This research is based upon work supported in part by the Office of the Director of National Intelligence (ODNI), Intelligence Advanced Research Projects Activity (IARPA), via [2022-21102100003]. The views and conclusions contained herein are those of the authors and should not be interpreted as necessarily representing the official policies, either expressed or implied, of ODNI, IARPA, or the U.S. Government. The U.S. Government is authorized to reproduce and distribute reprints for governmental purposes notwithstanding any copyright annotation therein.}}
\affil[1]{University of Colorado, Colorado Springs}
\affil[2]{Universidade Estadual de Campinas}
\affil[ ]{\tt\small wrobbins@uccs.edu, gabriel.bertocco@ic.unicamp.br, tboult@vast.uccs.edu}

\maketitle

\section{LFW-LD and CFP-LD Collection Settings}
\begin{figure}
    \centering
    \includegraphics[width=.9\linewidth]{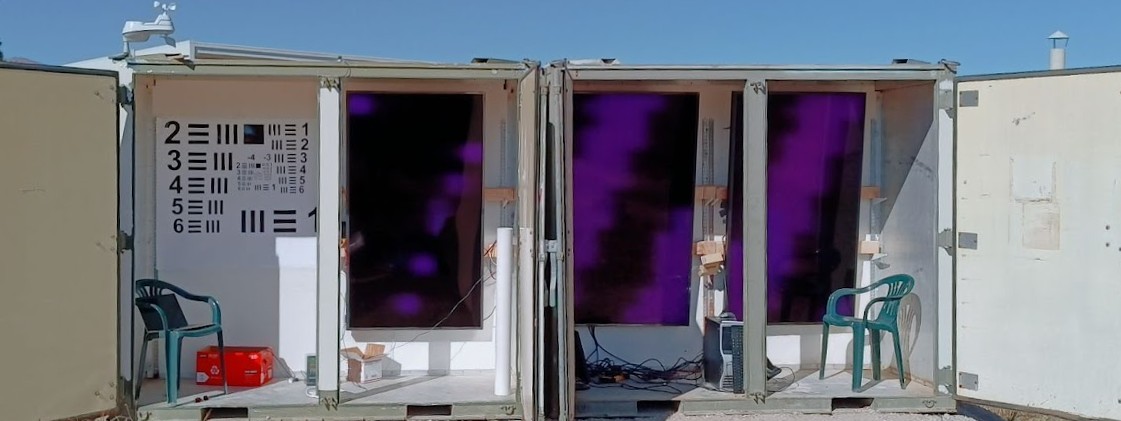}
    \caption{3x 75" 4k OLED 2,000 nit outdoor displays mounted in containers for recapture.}
    \label{fig:container}
\end{figure}

\begin{figure}
    \centering
    \includegraphics[width=.9\linewidth]{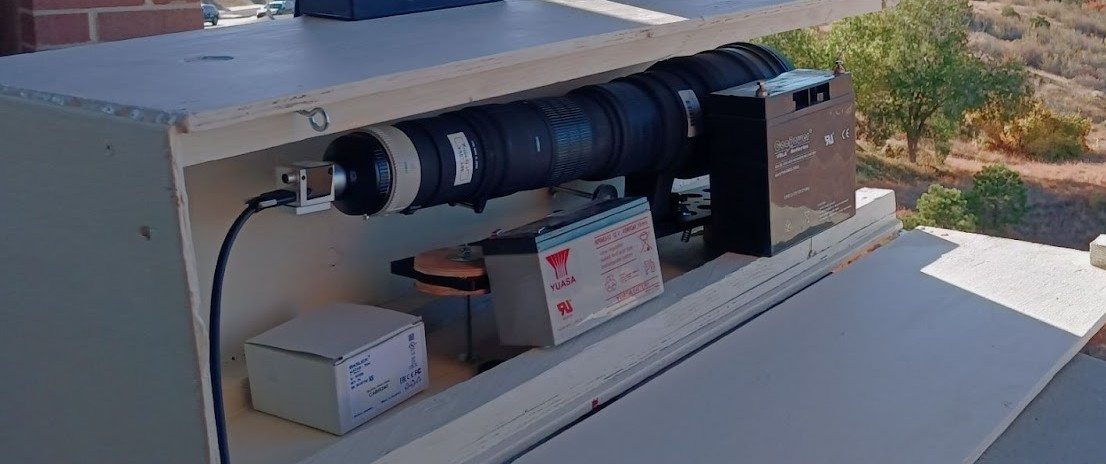}
    \caption{Lens and camera with custom mounting hardware for recapture.}
    \label{fig:camera}
\end{figure}

Specifications for the long-distance recapture datasets (``the LD datasets")  are briefly described in Section 4 of the main paper and are detailed below. The collection setup went through IRB approval and both he LFW and CFP dataset licenses allow
redistribution. Specifications of imaging equipment and collection conditions are shown in \tabref{specs}. \figref{container} shows the display and \figref{camera} shows the camera used for recapture. The LD datasets contain twelve recapture images for each display image as the capture occurs continuously over time and atmospheric turbulence is temporally variable (atmospheric effects are shown in Figure 3 of the main paper and \textit{in supplementary videos}). The nature of the data allows for research uses such as: frame selection, frame-aggregation, distortion robustness, quality prediction, and direct feature comparisons to the same image with and without real atmospheric turbulence.

To post-process the images, fixed regions from the screens are cropped, and then RetinaFace~\cite{retinaface} face detector is used to detect landmarks and re-align the images. Non-local mean denoising algorithm is used to reduce noise in the recapture images. \figref{samples} shows samples from the LD datasets. The recaptured videos are provided in the supplementary zip file, where atmospheric effects can be seen.
The collection process is ongoing, and further training and evaluation data for both face recognition and person re-identification will be released before CVPR. 
\begin{table}[H]
    \centering
    \begin{tabular}{p{3.7cm} l}
        \hline\hline
        Parameter & Value \\
        \hline
         Camera & Basler acA2440-35uc \\
         Lens focal length & 800mm +1.4x Extender\\
         Capture distance & 770 meters \\
         Integration time & 30$\mu$s \\
         Capture rate & 30 fps \\
         Wind speed & 5-15mph \\
         Temperature & 15$^{\circ} C$\\
         \hline\hline
    \end{tabular}
    \caption{Camera, weather condition, and display settings for the collection of LFW-LDand CFP-LD.\vspace{-10pt}}
    \label{tab:specs}
\end{table}

\setlength\dashlinedash{0.3pt}
\setlength\dashlinegap{1.5pt}
\setlength\arrayrulewidth{0.4pt}
\begin{figure*}
    \centering
    \includegraphics[width=.75\textwidth]{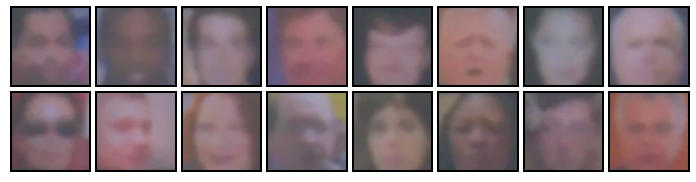}
    \caption{Sample images from the LD datasets. It can be seen that our recapture set-up yielded significant atmospheric turbulence effects (\textit{also see video provided with supplementary}). These datasets can facilitate research into 1) quality/confidence-aware models, 2) models that are robust to face-feature distortion, and 3) frame aggregation under atmospheric turbulence (12 frames are provided per display image).}
    \label{fig:samples}
\end{figure*}
\section{Face Recognition Adaptive Margin}
For face recognition, the AdaFace~\cite{Kim_2022_CVPR} loss is used, which uses an adaptive margin as a function of the feature norm. The adaptive margin includes both an angular margin $g_{angle}$ and an additive margin $g_{add}$ calculated as
\begin{equation}
    g_{angle} = -m\cdot\widehat{||x_i||},\; g_{add} = m\cdot\widehat{||x_i||} + m,
\end{equation}
where $\widehat{||x_i||}$ is the feature magnitude after normalizing the magnitudes with batch statistics. $m$ is a margin hyperparameter. The penalty for each sample can be represented with the piece-wise function $f$:
\begin{equation}
f(\theta_j, m) =  \begin{cases} 
      s\cos(\theta_j + g_{angle}) - g_{add} & j=y_i \\
      s\cos\theta_j & j\neq y_i \\ 
\end{cases}
\end{equation}
where $\theta_j$ is the angle between the feature vector from the backbone proxy class-center of the $j^{th}$ class. Scalar $s$ is a hyperparameter and $y_i$ is ground truth. The final AdaFace loss $\mathcal{L}_{adaface}$ is then calculated as follows:
\begin{equation}
     \mathcal{L}_{adaface} =  -\frac{1}{N}\sum_{i=1}^{N}{\log\frac{e^{f(\theta_j, m))}}{e^{f(\theta_j, m))} + \sum_{j\neq y_i}{e^{f(\theta_j, m)}}}}.
\end{equation}

\section{Proxies and centers definitions for PReID}

Here we present how we calculate the class proxies introduced in Section 3.2 in the main paper. Without loss of generality, consider a class $C = \{c_{1}, ..., c_{N_{C}}\}$ in the dataset with $N_{C}$ examples. To calculate the proxies set, we start by randomly selecting a sample $c_{i} \in C$ ($ 1 \leq i \leq N_{C}$) to be the first proxy, and we calculate the distance between $c_{i}$ and each element in $C$ and store these distances in a cumulative vector $V_{C} \in R^{N_{C}}$. We call the first proxy as $p_{C}^{1} = c_{i}$ .To calculate the second proxy, we consider the element with  the furthest distance to the first proxy (the sample with maximum distance value in $V_{c}$). Formally:

\begin{equation}
\label{eq:second_proxy_calculation}
    p_{C}^{2} :=  \argmax V_{C}  
\end{equation}

After that, we calculate the distance of $p_{C}^2$ to all samples in $C$ to obtain the distance vector $D(p_{C}^2) \in R^{N_{c}}$. Then we update $V_{C}$ considering its current values (the distances of the class samples to the first proxy) and $D(p_{C}^2)$ (the distance of the class samples to the second proxy) following the formulation:

\begin{equation}
\label{eq:second_cumulative_vectors_updating}
    V_{C} := min(V_{C}, D(p_{C}^{2}))  
\end{equation}
where $min(.,.)$ is the element-wise minimum operation between two vectors. More specifically, the $j^{th}$ position of $V_{C}$ will hold the minimum distance of the sample $c_{j} \in C$ considering the first and second proxies. So the $j^{th}$ position holds the distance of $c_{j}$ to the closest proxy, and the maximum value in $V_{C}$ is from the sample most apart from both proxies. We consider this sample as the next proxy $p_{C}^3$. To obtain $p_{C}^3$, we apply again Eq.~\ref{eq:second_proxy_calculation} but considering the updated $V_{C}$ calculated from Eq.~\ref{eq:second_cumulative_vectors_updating}, and repeat the whole process again for the new proxy. We write both equations in their general formats:

\begin{equation}
\label{eq:next_proxy_calculation}
    p_{C}^{t} :=  \argmax V_{C}^{t-1} 
\end{equation}

\begin{equation}
\label{eq:cumulative_vectors_updating}
    V_{C}^{t} := min(V_{C}^{t-1}, D(p_{C}^{t}))  
\end{equation}

As explained before, we initialize $V_{C}^{1} := D(p_{C}^{1})$ where $p_{C}^{1}$ has been randomly selected from $C$ to be the first proxy. We keep alternating between Equations~\ref{eq:next_proxy_calculation} and~\ref{eq:cumulative_vectors_updating} until $t = 5$ to get five proxies per class. During training, for a sample $X_{i} \in B$ (where $B$ is the batch), we call $P_{i}$ by the proxies set of its class and $N_{i}$ by the set of the top-50 closest negative proxies and use them to calculate $\mathcal{L}_{proxy}$ in Eq. 3 of the main paper. After that, $\mathcal{L}_{proxy}$ loss is employed along with on $\mathcal{L}_{center}$ in Eq. 4 in the main paper. The class proxy calculation is used just for PReID training.  

\begin{figure*}
    \centering
    \includegraphics[width=.7\textwidth]{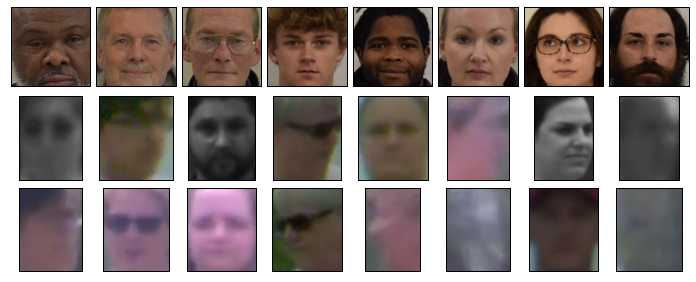}
    \caption{Samples from the gallery set (top row) and query set (bottom two rows) from the long-range dataset (LRD) used in the main paper. Query images are taken at distances between 100-500 meters. All subjects consented to image use in publication.}
    \label{fig:lrd}
\end{figure*}
\begin{table*}[]
    \centering
    \small
    \begin{tabular}{l l l  l }
    \multicolumn{4}{l}{\large\textbf{Dataset Reference}} \\
    \hline\hline
        Dataset & Modality & Evaluation Metric & Characteristics \\
        \hline
        CFP-LD & face & 1:1 verification  & Recapture dataset at 770m; strong atmospheric turbulence\\
        LFW-LD & face & 1:1 verification  & Recapture dataset at 770m; strong atmospheric turbulence\\
        \hdashline
        LRD    & face & R-1, R-5, TPIR@FPIR=1e-1,1e-2 & HQ gallery images; query images up to 500m. Government-use.  \\
        \hdashline
        CFP-FP & face & 1:1 verification & Relatively high-quality; frontal-profile pairs \\
        LFW & face & 1:1 verification & Relatively high-quality\\
        AgeDB-30 & face & 1:1 verification & Relatively high-quality; pairs with 30 year difference \\
        IJB-C & face & TAR@FAR=1e-4 & Mixed-quality\\
        IJB-S & face & R-1, R-5, TPIR@FPIR=1e-1,1e-2  & High-quality gallery; low spatial resolution faces in probe video\\
        TinyFace & face&  R-1, R-5 & Low spatial resolution probe and gallery \\
        \hline
        DeepChange & PReID & mAP, R-1 & 16-cameras low-resolution with 450 clothes-changing identities\\
        Market & PReID & mAP, R-1 & 6-cameras low/high-resolution with 751 same-clothes identities\\
        MSMT17 & PReID & mAP, R-1 & 15-cameras low/high-resolution with 1041 same-clothes identities\\
        \hline\hline
        
    \end{tabular}
    \caption{Reference table of datasets used in the main paper. The CFP-LD and LFW-LD datasets are proposed in Section 4 of the main paper, the LRD (long-range dataset) is a government-use dataset, and all other datasets are from prior works. For CFP-LD and LFW-LD, we use the same evaluation as standard LFW and CFP. For LRD, we use the same metrics as for IJB-S because the dataset has the same gallery/query format. All other evaluation metrics follow standard practice from prior works. Mean Average Precision (mAP), Rank-1 (R-1), and Rank-5 (R-5) are retrieval metrics.}
    \label{tab:datasets}
\end{table*}

\section{Additional Implementation details for PReID}

To train the clean and distortion models, we employ the Adam~\cite{kingma2014adam} optimizer with weight decay of $5e^{-4}$ and initial learning rate of $3.5e^{-4}$. We train both models for $250$ epochs and divide the learning rate by $10$ every $100$ epochs. As explained, the number of proxies per class is fixed in 5 (i.e., $\forall_{i} |P_{i}| = 5$) for all datasets. To create the batch to optimize the clean model, we adopt a similar approach to the PK batch strategy~\cite{hermans2017defense} in each we randomly choose $P$ identities and, for each identity, $K$ clean images (without distortion). To train the distortion model, we sample $K$ clean images, and $K$ distorted images randomly sampled from five different levels of distortion strength. We also apply Random Crop, Random Horizontal Flipping, Random Erasing, and random changes in the brightness, contrast, and saturation as data augmentation. 

To improve the performance, we adopt the Mean-Teacher~\cite{tarvainen2017mean} to self-ensemble the weights of the backbones along the training. Considering both Clean and Domain-Adaptive backbones with parameters $\theta_{cl}$ and $\theta_{da}$ (which are initialized with weights pre-trained on Imagenet), respectively, we keep another backbone for each one with parameters $\Theta_{cl}$ and $\Theta_{da}$ with the same architecture to self-ensemble their weights along training through the following formula:

\begin{equation}
\label{eq:mean_teacher}
    \Theta_{s}^{t+1} := \beta\Theta_{s}^{t} + (1-\beta)\theta_{s}^{t}  
\end{equation}

\noindent where $s \in \{cl, da\}$, $\beta$ is a hyperparameter to control the inertia of the weights, and $t$ is the instant of time. We set $\beta = 0.999$ for all models following prior PReID works~\cite{ge2020mutual, zhai2020multiple}. We use the backbones $\Theta_{cl}$ and $\Theta_{dl}$ for the final evaluation. 

\section{Dataset Reference}
Many datasets from two different modalities are used in the main paper. \tabref{datasets} is provided as a reference for the different characteristics of the datasets. \figref{samples} shows samples from the LD datasets and \figref{lrd} shows samples from the government-use long-range-dataset.

\section{IJB-S Evaluation Details}
The IJB-S~\cite{ijbs} is a surveillance dataset that is distributed as a set of gallery images for 202 identities and over 30 hours of query videos. The dataset has 15 million face bounding-box annotations. To process the data, we follow the following steps:
\begin{enumerate}
    \item Extract all 15 million annotated face regions from all images and videos.
    \item Run all extracted regions through MTCNN~\cite{mtcnn} face detector. MTCNN detected 7.28M/15M face regions.
    \item Use face landmarks from MTCNN for an affine transformation to fixed positions on 112x112 image --- zero-padding is added if necessary.
\end{enumerate}
Evaluation is performed with the surveillance-to-booking and surveillance-to-surveillance protocols. Surveillance-to-booking protocols uses videos with thousands of frames for a query and a template of multi-view high-quality gallery images. Surveillance-to-surveillance uses surveillance video for both the probe and the gallery. Seven gallery images are used for each of 202 identities and 7,287,724 query face detections are used.

\section{Further Ablation Studies}
\textit{Distortion Augmentation.} In the main paper, we propose the use distortion augmentation inspired by atmospheric turbulence. \tabref{ablation_augmentation} shows a comparison to a combination of other similar augmentations that have been used in computer vision: down-sampling and Gaussian blur. Gaussian blur and down-sampling are applied at equally challenging levels as the distortion augmentation (as measured by the loss). In \tabref{ablation_augmentation} it can be seen that distortion augmentation performs better than Gaussian blur and down-sampling on both face recognition and person re-identification benchmarks.
\begin{table}[ht]
\footnotesize
\centering
\resizebox{1.0\linewidth}{!}{\begin{tabular}{l c c  c c c c }
\hline\hline
\multicolumn{1}{c}{} &
\multicolumn{2}{c}{IJB-C} &
\multicolumn{2}{c}{CFP-LD} &
\multicolumn{2}{c}{TinyFace} \\ \hline
DS+GB & \multicolumn{2}{c}{96.48} & \multicolumn{2}{c}{77.13} & \multicolumn{2}{c}{73.39} \\
Distortion Aug & \multicolumn{2}{c}{\textbf{96.91}} & \multicolumn{2}{c}{\textbf{78.16}} & \multicolumn{2}{c}{7\textbf{4.11}} \\ [1.5ex] \hline 
\multicolumn{1}{c}{} &
\multicolumn{2}{c}{Market} &
\multicolumn{2}{c}{MSMT17} &
\multicolumn{2}{c}{DeepChange} \\ \hline
& mAP & R1 & mAP & R1 & mAP & R1 \\ \hline
DS+GB & 78.0 & 91.2 & 44.7 & 69.5 & 16.2 & 51.5\\
Distortion Aug & \textbf{86.3} & \textbf{94.7} & \textbf{55.4} & \textbf{78.5} & \textbf{20.2} & \textbf{58.6} \\ \hline
\multicolumn{7}{l}{DS+GB=down-sampling + Gaussian blur}
\end{tabular}}
\caption{A comparison between training augmentations. The distortion augmentation performs better than using Gaussian blur and down-sampling.}
\label{tab:ablation_augmentation}
\end{table}

\textit{Feature fusion methods.} Our DaliID method uses a magnitude-weighted fusion of features from two backbones (see Figure 2 of the main paper). We also performed experiments with learned fusion layers. As shown in \tabref{fusion}, we found that the magnitude-weighted fusion outperformed learned fusions.
\begin{table}[h]
    \centering
    \small
    \begin{tabular}{l|c c c}
        \hline
        Fusion & IJB-C & CFP-LD & TinyFace \\
        \hline
        magnitude weighted fusion & \textbf{97.40} & \textbf{78.97} & \textbf{73.98} \\
        linear layer & 97.07 & 78.19 & 73.87  \\
        attention layer & 97.20 & 78.26 & 73.84 \\
        transformer decoder & 97.22 & 77.98 & 73.82 \\
        \hline
    \end{tabular}
    \caption{Experiments with three different learning methods to combine the feature vectors from the clean and distortion-adaptive backbones. Perhaps surprisingly, we get the best results without learning a final representation but rather performing magnitude-weighted fusion.}
    \label{tab:fusion}
\end{table}

\begin{figure}[ht]
\centering
\subfloat[Impact of different $\lambda$ values]{\includegraphics[width=1.5in]{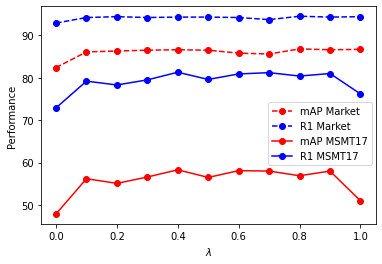}
\label{fig:lambda_impact}}
\hfil
\subfloat[Impact of different $\tau$ values]{\includegraphics[width=1.5in]{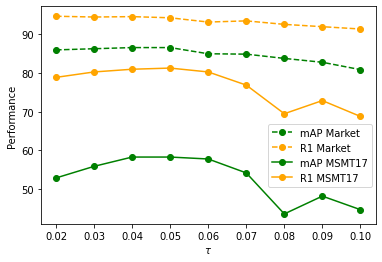}
\label{fig:tau_impact}} \\
\caption{Analysis of impact of the parameters $\tau$ and $\lambda$ on the final loss function considering the training of the Distortion-Adaptive Backbone.}
\label{fig:lambda_tau_impact}
\end{figure}

\textit{PReID Parameter Analysis}. There are two hyper-parameters on the final loss function (Eq. 4 in the main paper) for Person Re-Identification: $\tau$ value to control the sharpening of the probability distribution in its both terms, and $\lambda$ value to weight the contribution of $\mathcal{L}_{proxy}$ term. The impact of these parameters on the performance of the Distortion-Adaptive Backbone is shown in Figure~\ref{fig:lambda_tau_impact}. 

For $\lambda$ in Figure~\ref{fig:lambda_impact} we see stable performance for Market along different values after $\lambda = 0.1$, while for MSMT17 we see a peak at $\lambda = 0.4$, then a suitable decrease after this value. For both datasets, we see a performance drop for $\lambda = 0.0$ (no $\mathcal{L}_{proxy}$), showing the proxy-based loss term has a positive impact on training while an equal contribution of both terms $\lambda = 1.0$ hurts the performance mainly for MSMT17. Since MSMT17 is more challenging, we select $\lambda = 0.4$ as the operational value. Further analysis of the impact of $\mathcal{L}_{proxy}$ is presented in Table~\ref{tab:ablation_weighting}.

The impact of $\tau$ is shown on Figure~\ref{fig:tau_impact}. The performance drops when $\tau$ is lower than $0.04$ for MSMT17 but a stable behavior for Market, while values greater than $0.06$ deteriorate the performance for both datasets. To achieve a good trade-off considering the dataset complexities, we choose $\tau = 0.05$.

\textit{Impact of pooling operations in evaluation (PReID)}. As shown in Figure 2 in the main paper, the Evaluation is performed by a weighted combination of the decisions from Clean and Distortion-Adaptive backbones. The weights $W_{clean}$ and $W_{distortion}$ are the maximum magnitudes of the feature vectors for each query and gallery image pair for each backbone. Among the different pooling strategies, we choose Global Average Pooling (GAP), Global Max Pooling (GMP), and a combination of both (GAP+GMP) to check the impact on final performance. The performances are reported in Table~\ref{tab:ablation_weighting}, which is an extension of Table 6 from the main paper. Note that in this case, the pooling operations are \textbf{just to calculate the magnitudes}, since the final representation, as explained in section 5.2 of the main paper, is always obtained by the element-wise sum of the output of the GAP and GMP layers for PReID. 

\begin{table}[ht]
\footnotesize
\centering
\resizebox{1.0\linewidth}{!}{\begin{tabular}{l | c c | c c | c c }
\hline\hline
\multicolumn{1}{c|}{} &
\multicolumn{2}{|c|}{Market} &
\multicolumn{2}{|c}{MSMT17}&
\multicolumn{2}{|c}{DeepChange} \\ \hline
 Setup & mAP & R1 & mAP & R1 & mAP & R1 \\ \hline
Baseline ($\theta_{cl}$) & 86.6 & 94.2 & 57.6 & 80.3 & 20.5 & 59.3\\
Distortion Aug & 86.3 & \textbf{94.7} & 55.4 & 78.5 & 20.2& 58.6 \\
Distortion-Adaptive (no $\mathcal{L}_{proxy}$) & 82.4 & 92.9 & 47.9 & 72.9 & 19.2 & 55.6\\
Distortion-Adaptive ($\theta_{da}$)& 86.6 & 94.3 & 58.3 & 81.3 & 20.7 & 59.2 \\ \hline
GMP & 87.6 & 94.4 & 60.5 & 82.1 & 21.9 & 60.7\\ 
GMP+GAP & 87.6 & 94.4 & 60.6 & 82.1 & 21.8 & 60.8\\ \hline
\textbf{DaliReID (GAP)} & \textbf{87.6} & 94.5 & \textbf{60.6} & \textbf{82.1} & \textbf{21.9} & \textbf{60.8} \\
\hline\hline
\end{tabular}}
\caption{Extension of Table 6 of the main paper for Person Re-Identification. We consider different strategies to combine the features from Clean and Distortion-Adaptive Backbones.}
\label{tab:ablation_weighting}
\end{table}

We see among GAP, GMP, and GAP+GMP, we have a similar performance in evaluation, with a slighter improvement for GAP. All of them have similar performances over the final result showing our proposed fusion strategy is robust to different pooling operations.

\section*{Code and Data Release}
Code will be made publicly available upon acceptance. The LD datasets will be made available for academic use upon acceptance.

\balance
{\small
\bibliographystyle{ieee_fullname}
\bibliography{refs}
}